\documentclass[conference]{IEEEtran}
\IEEEoverridecommandlockouts
\usepackage[switch]{lineno}
\usepackage{textcomp}
\usepackage{cite}
\usepackage{amsmath,amssymb,amsfonts}
\usepackage{algorithm}
\usepackage{algorithmic}
\usepackage{graphicx}
\usepackage{url}
\usepackage{verbatim}
\usepackage{comment}
\usepackage[a4paper, total={184mm,239mm}]{geometry}
\def\BibTeX{{\rm B\kern-.05em{\sc i\kern-.025em b}\kern-.08em
    T\kern-.1667em\lower.7ex\hbox{E}\kern-.125emX}}
\begin{document}
\title{Towards Visual Syntactical Understanding} 

\author{\IEEEauthorblockN{Sayeed Shafayet Chowdhury, Soumyadeep Chandra, and Kaushik Roy}
\IEEEauthorblockA{\textit{Elmore Family School of Electrical and Computer Engineering, }
\textit{Purdue University, }
West Lafayette, IN 47907, USA\\
chowdh23@purdue.edu, chand133@purdue.edu, kaushik@purdue.edu}
}
\maketitle

\begin{abstract}
Syntax is usually studied in the realm of linguistics and refers to the arrangement of words in a sentence. Similarly, an image can be considered as a visual `sentence', with the semantic parts of the image acting as `words'. While visual syntactic understanding occurs naturally to humans, it is interesting to explore whether deep neural networks (DNNs) are equipped with such 
reasoning.~To that end, we alter the syntax of natural images (e.g.~swapping the eye and nose of a face), referred to as `incorrect' images, to investigate the sensitivity of DNNs to such syntactic anomaly.~Through our experiments, we discover an intriguing property of DNNs where we observe that state-of-the-art convolutional neural networks, as well as vision transformers, fail to discriminate between syntactically correct and incorrect images when trained on only correct ones. 
~To counter this issue and enable visual syntactic understanding with DNNs, we propose a three-stage framework- (i) the `words' (or the sub-features) in the image are detected, (ii) the detected words are sequentially masked and reconstructed using an autoencoder, (iii) the original and reconstructed parts are compared at each location to determine syntactic correctness. The reconstruction module is trained with BERT-like masked autoencoding for images, with the motivation to leverage language model inspired training to better capture the syntax. Note, our proposed approach is unsupervised in the sense that the incorrect images are only used during testing and the correct versus incorrect labels are never used for training.~We perform experiments on CelebA, and AFHQ datasets and obtain classification accuracy of 92.10\%, and 90.89\%, respectively. Notably, the approach generalizes well to ImageNet samples which share common classes with CelebA and AFHQ without explicitly training on them. 
\end{abstract}


\section{Introduction}
\label{sec:introduction}
Semantics and syntax are two widely studied concepts for languages.~Usually, semantics refers to the holistic meaning of sentences, whereas syntax contains the set of rules defining the proper order of words 
\cite{SyntaxvsSemantics}.~The notion of semantics and syntax can be similarly applied to visual understanding  \cite{tang1979syntactic}.~In the pre-deep learning (DL) era, both syntactic \cite{fu1974syntactic} and semantic \cite{baird1973paradigm} approaches had been adopted for image analysis. However, with the advent of deep neural networks (DNNs), convolutional neural networks (CNNs), and recently, Vision Transformers (ViTs) have become the {\em de facto} models for vision. While the semantic understanding (e.g.~classification, object detection, segmentation) of these DNNs has been studied rigorously \cite{krizhevsky2012imagenet,long2015fully}, their syntactic reasoning capabilities remain to be investigated in detail.

Motivated by the syntactic understanding of natural language processing (NLP) models \cite{kulmizev2020neural,targeted}, we argue that images can be perceived to have their own syntax as well. If we consider a visual language, an image can be thought of as a  \textit{`sentence'}, where the semantic parts of the image form the \textit{`words'}.~For example, for an image of a face (sentence), the constituent words can be eye, ear, nose, mouth, etc. However, a random spatial combination of these will not result in a meaningful face. Rather, for an image to be syntactically correct, the parts need to conform to a specific arrangement, which we refer to as \textit{`Image Syntax'}. In terms of a face, a proper syntax for a row-major scan could be eyes, followed by left ear, nose, right ear, and mouth. 
Such syntactic understanding is innate to humans; however, it is worth investigating whether DNNs possess similar syntactic reasoning. 

\begin{figure*}[t]
  \centering
\includegraphics[width=0.95\linewidth]{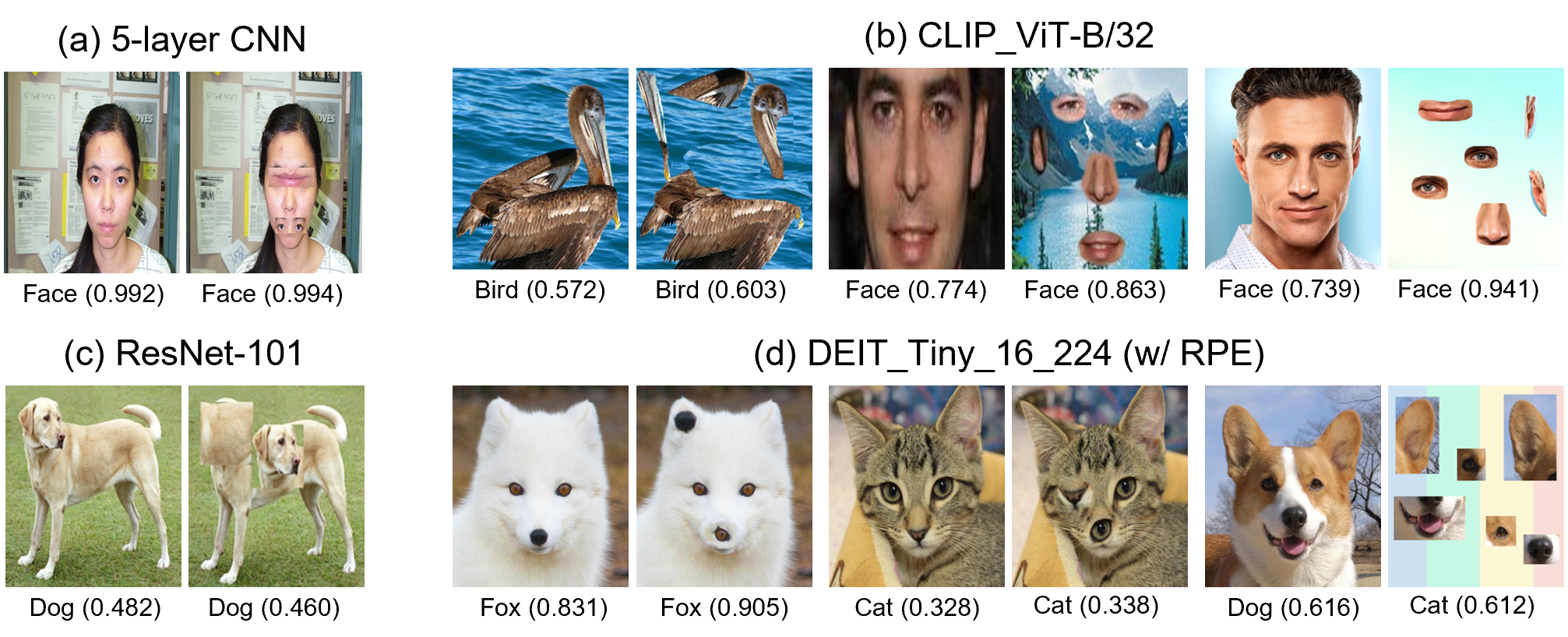}
 
   \caption{Predictions on syntactically correct and incorrect images using (a) 5 layer CNN, (b) CLIP$\textunderscore$ViT-B/32, (c) ResNet-101, (d) DEIT$\textunderscore$Tiny$\textunderscore$16$\textunderscore$224 (with relative positional encoding). For each pair, the correct image is on the left with the corresponding incorrect one on the right. The prediction probabilities are shown in parentheses with the predicted class.}
  \label{fig:motivation}

\end{figure*}

To this end, we artificially manipulate natural images to alter their syntax as shown in Fig.~\ref{fig:motivation} (e.g.~swapping different parts of an image, separating semantic face parts, and distributing them on a random background). Subsequently, these incorrect images (II) are classified with state-of-the-art (SOTA) CNNs and ViTs along with the corresponding correct images (CI). Interestingly, we observe that the DNNs fail to discriminate between syntactically CI and IIs. Moreover, the IIs often receive higher prediction probability compared to the CIs, contrary to our expectations. Thus, we reveal an interesting failure mode of DNNs in the context of visual syntactic reasoning. Notably, ViT-based models also exhibit this property despite having positional encoding of image patches.

Having observed the above mentioned property of DNNs, we take a step forward to design a method to enable detection of CI versus II. 
To instill syntactic comprehension within a DL pipeline, we propose a three-stage modular approach. First, the constituent words (parts) of the image are detected. The set of object parts (like the vocabulary in a language) is pre-defined and this step localizes the parts present in the input irrespective of their syntax.~Next, we mask the detected parts one at a time and use a masked autoencoder \cite{mae} to reconstruct them.~The reconstruction module is trained by random masking during training. We hypothesize that by learning to reconstruct the masked words, the model will learn their correct spatial configuration since they are only exposed to images with proper syntax during training. This approach is motivated by pre-training of bidirectional encoder representations from transformers (BERT) \cite{bert}, where the language model is trained by masking words and learning to reconstruct them. Our intuition is that mimicking the training of a language model would improve the syntactic understanding of vision models. ~Finally, the given and the reconstructed images are compared at each location to establish the overall syntactic correctness.~Some salient features of the proposed method are- (i) it is unsupervised as the IIs are only used for testing, (ii) in addition to the binary correct versus incorrect classification, the method provides an interpretation of what part of the image is wrong (for incorrect inputs), (iii) the reconstructed output provides the correct syntactic configuration. The proposed method is evaluated on CelebA and AFHQ and we obtain classification accuracy of 92.10\%, and 90.89\%, respectively. Notably, our approach generalizes well to ImageNet and Caltech 101 samples which share common classes with CelebA and AFHQ, even when the model is not trained on them. 
 
\begin{figure*}[t]
  \centering
\includegraphics[width=0.95\linewidth]{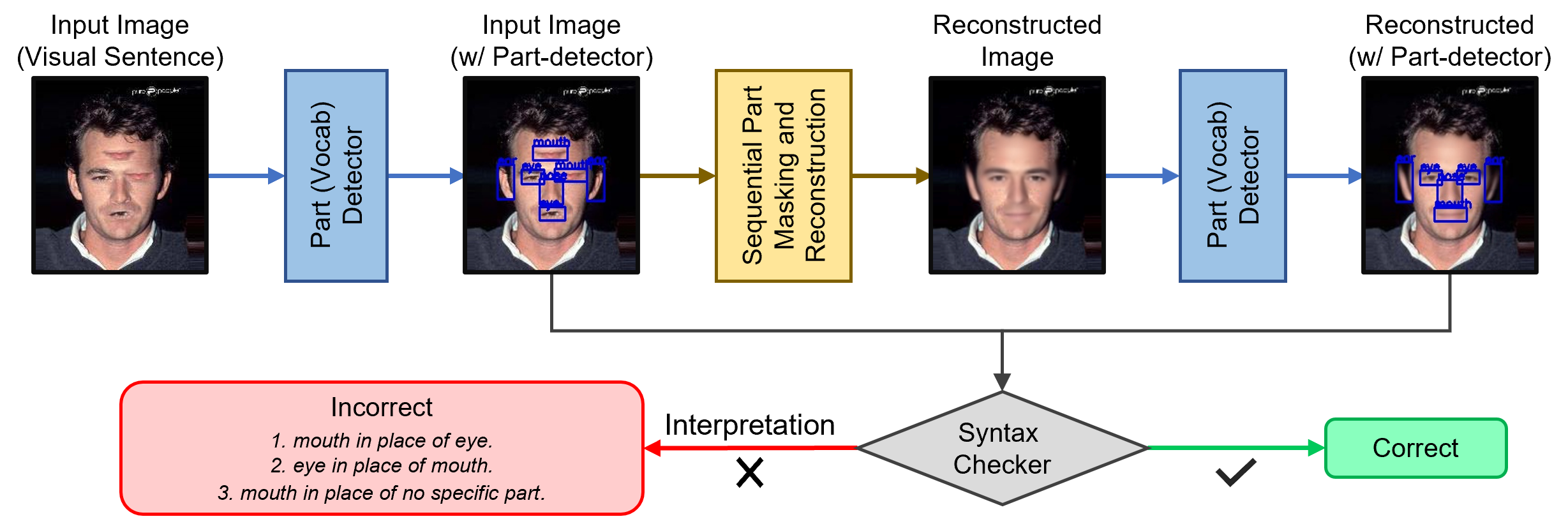}
 
   \caption{Schematic of the proposed method. First, the input image is passed through a part detector (PD), and each detected word (part) is then sequentially masked and reconstructed. The words present in the reconstructed image are then detected using the same PD. Finally, a syntax checker compares the original and reconstructed parts at each location and evaluates syntactic correctness. Additionally, for incorrect inputs, interpretation is provided of \textit{what} is incorrect.}
  \label{fig:method}

\end{figure*}
To summarize, the main contributions of this work are-

\begin{itemize}
\setlength{\itemindent}{0em}

\item To the best of our knowledge, this is the first work that introduces the aspect of visual syntactic understanding in the context of DNNs. Notably, we discover that current DL models do not possess such understanding inherently.

\item We propose an unsupervised technique inspired by BERT-like masked autoencoding which can successfully evaluate syntactic correctness. 

\item Our proposed method is interpretable as it provides explanations of syntactic anomaly. Moreover, the output reconstructs a feasible version of the expected input.

\item  This approach can be considered as part of a neuro-symbolic pipeline, where the word detection and reconstruction modules act as neural blocks and syntax checking forms the symbolic part. 

\end{itemize}

\section{Related Works}
\label{related_works}
\subsection{Syntactic Analysis of LMs} 
Syntactic Analysis of natural language processing (NLP)-based language models (LMs) aims to evaluate them with two closely matched sentences differing only in their grammar. The authors in \cite{targeted} proposed that the probability of the grammatically correct sentence should be higher than the probability of the grammatically incorrect one, but the studies are mostly limited to subject-verb agreement and long short term memory (LSTM) models. A different approach is to decode linguistic properties like part-of-speech, and named-entity recognition by probing the hidden states  \cite{tenney2019you,peters2018dissecting} of models such as BERT \cite{bert}.~A related work \cite{kulmizev2020neural} proposed a probe for extracting directed dependency trees to BERT and ELMo\cite{peters1802deep} trained on multiple languages.~Warstadt et al.~\cite{blimp} introduced a benchmark of linguistic minimal pairs to evaluate LMs on major grammatical phenomena. Such targeted syntactic study of LMs was further refined in \cite{refining} through weighted evaluation.~However, a similar syntactic assessment has not been performed for DL-based vision models. Note, the study of syntax for images is inherently more complex compared to a language as- (i) due to occlusion, pose variations, crop etc.,~only a portion of an image may be visible (and still be syntactically correct), whereas, a part of a sentence (such as \textit{`I am a'} or \textit{`am a person'}) may not be meaningful. (ii) There is no left-to-right ordering in images as in language, so language grammar is essentially 1D, while image grammar is 2D \cite{zhu2007stochastic}. (iii) Unlike language, in vision models, objects can appear at arbitrary scales.

\subsection{Image Grammar}~A semantic-syntactic approach to image generation was proposed in \cite{tang1979syntactic}. Fu et al.~\cite{fu1974syntactic} studied scene understanding using rule-based hierarchical parse graph representation.~The authors in \cite{zhu1996forms} introduced a shape grammar for shape matching and recognition.~Grenander \cite{grenander1993general} proposed pattern matching on a set of graphs based on primitives which have multiple attributes and connect with other elements.~A similar stochastic grammar of images was studied in \cite{zhu2007stochastic}, where an image was decomposed in a hierarchical and-or parse graph. However, these works mostly relied on simple primitives based on simplistic data and represent the pre-DL era.~In contrast, we focus on the syntactic understanding of vision-based modern  DNNs on complex data.     

\subsection{Image Inpainting} Inpainting involves masking parts of the input and learning to recover them.~Pre-training of BERT \cite{bert} masks out words and the model learns to reconstruct them.~Similar approaches have been adopted for images in \cite{beit,mae}.~The masking and reconstruction module in our work utilizes masked autoencoding \cite{mae}.~Note, these works mostly focus on generation by random masking and are completely unrelated to syntax consideration.~On the contrary, the proposed reconstruction module specifically masks words (parts) of the image for subsequent syntactic evaluation.    

\subsection{Anomaly Detection} Anomaly detection methods usually consider anomalous samples to belong to a different class or distribution compared to the normal class. One-class support vector machine (OC-SVM) is a classic kernel-based method for anomaly detection \cite{scholkopf1999support}. Another related approach is one-class deep support vector data description \cite{svdd}.  Deep energy-based model \cite{dsebm} can also be used for image anomaly detection where high energy samples are considered anomalous. Other unsupervised anomaly detection methods involve generative adversarial networks (GANs) \cite{anogan,ganomaly,ADGAN}, kernel density estimation \cite{kde} etc. 
However, unlike the assumption of these approaches, syntactic anomaly does not alter the underlying class. As such, current detection schemes are not suitable for syntax checking, even though erroneous syntax can be considered a form of image anomaly. This claim is validated in the results section \ref{comparison}.

\section{An Intriguing Property of SOTA DNNs -- Lack of Visual Syntactic Understanding}
Syntactic understanding is a thoroughly analyzed topic for language models \cite{kulmizev2020neural,targeted}. To demonstrate the effect of input syntax on language models, we experiment with the state-of-the-art (SOTA) T5 transformer for English-to-German translation. In this case, we obtain the following results (input followed by translation in brackets)- `In the beginning, we liked the game' (\textit{Am Anfang hat uns das Spiel gefallen}), and `In the we game beginning liked the' (\textit{Im Spiel begann es, die}). Clearly, the syntax of the prepositional phrase (`In the beginning') impacts the output. Similarly, high performance across verb phrases, prepositional phrases, subject/object clauses, reflexive pronouns, and negative polarity items were reported in \cite{targeted,refining}, showing NLP models' preference for syntactic formalism. Taking inspiration from such syntactic understanding of NLP-based language models, in this section, we pose the question: \textit{Do SOTA CNNs and ViTs show similar syntactic understanding in the context of vision?}  

Before delving deeper into our investigation, let us set the stage with some definitions. We consider an image as a \textit{`visual sentence'} which is composed of a set of semantic parts, which we consider to be the \textit{`words'} of that visual sentence (image). For faces, this set comprises eyes, ears, nose, and mouth; for airplanes, the set may contain cockpit, wings, body, and tail, and likewise for other classes. However, a random spatial combination of these words may not result in a meaningful image. For example, let us consider the last pair of images of the top row in Fig.~\ref{fig:motivation}. While the first image depicts an actual face, the 2$^{nd}$ (rightmost) one just contains the words (face parts) in random order. As humans we can still discern it originates from a face, the syntactic anomaly is clearly understandable. Similarly, the 1$^{st}$ pair of faces in the top row of Fig.~\ref{fig:motivation} represents another visual syntactic distortion (with eyes and mouth swapped) and so on. 

Our goal is to understand the effect of such syntactically distorted images on the prediction of today's DNNs. With that objective, we artificially manipulate the syntax of natural images as shown in Fig.~\ref{fig:motivation}. Following this, we subject the syntactically incorrect images (IIs) to classification using SOTA CNNs and ViTs alongside the corresponding correct images (CIs). Intriguingly, we note that these DNNs struggle to differentiate between the CIs and IIs.~As illustrated in Fig.~\ref{fig:motivation}(a), upon receiving a fake face input (eyes and mouth swapped), a 5-layer CNN (trained to classify between faces and bikes) predicts the II as a face with an even higher probability compared to the CI. Similarly, ResNet-101 (Fig.~\ref{fig:motivation}(c)), when trained on ImageNet, classifies an incorrect dog image with similar confidence as the original. We obtain similar results with a ViT-B/32 CLIP model \cite{clip} as shown in Fig.~\ref{fig:motivation}(b).~Note, even when the different face parts are scattered on a random background, the prediction probabilities for face class (0.863 and 0.941) increase compared to an actual face (0.774 and 0.739, respectively), contrary to our expectation. These results are particularly surprising since ViTs despite having positional encodings are unable to distinguish between CI versus II. Furthermore, we perform similar experiments on ViTs with relative positional encoding (RPE) \cite{rpevit}, as depicted in Fig.~\ref{fig:motivation}(d). Surprisingly, even with RPE, these SOTA models remain insensitive to erroneous syntactic configurations. Even in this case, the IIs often receive higher prediction probability compared to the corresponding CI. Note, that for all cases, the models were trained on CIs only. These results demonstrate a fascinating counter-intuitive property of current DNN-based vision models -- they do not inherently capture syntactic understanding like the NLP models \cite{kulmizev2020neural,refining} do. While the discovery of this novel failure mode of DNN-based vision models is captivating, we also propose a potential approach to enable visual syntactic understanding in these models, as outlined in the following.

\section{Proposed Methodology}
\label{proposed_method}
The proposal of a DNN pipeline with visual syntactic reasoning capabilities needs to consider some desired characteristics. First, the syntax should be captured in an unsupervised manner as- (a) the number of possible incorrect syntactic configurations is extremely large, and hence, it is prohibitive to train on such data, (b) unsupervised method is more human-like as we can detect visual syntactic anomalies without ever seeing them earlier. Second, the method should be able to explain what is incorrect if the input contradicts the correct syntax. Lastly, for incorrect images, an approximation of the correct version should be provided. Keeping these in mind, we present a pipeline with 3 salient blocks as shown in Fig.~\ref{fig:method}, the details of which are described next.

\subsection{Semantic Part (Vocabulary) Detector}
To understand the syntax, it is crucial to first recognize the vocabulary. If the words of a language are unknown, it is infeasible to comprehend the proper syntax. As such, for any class of image or \textit{`visual sentence'}, a set of semantic parts needs to be specified as the \textit{`words'} of that class. To detect these words in a sentence, we train a semantic part detector (PD) by fine-tuning a faster-RCNN \cite{faster} having a ResNet-50 backbone with feature pyramid network. Note, only CIs are used for training.   
 \begin{figure}[t]
  \centering
\includegraphics[width=0.95\linewidth]{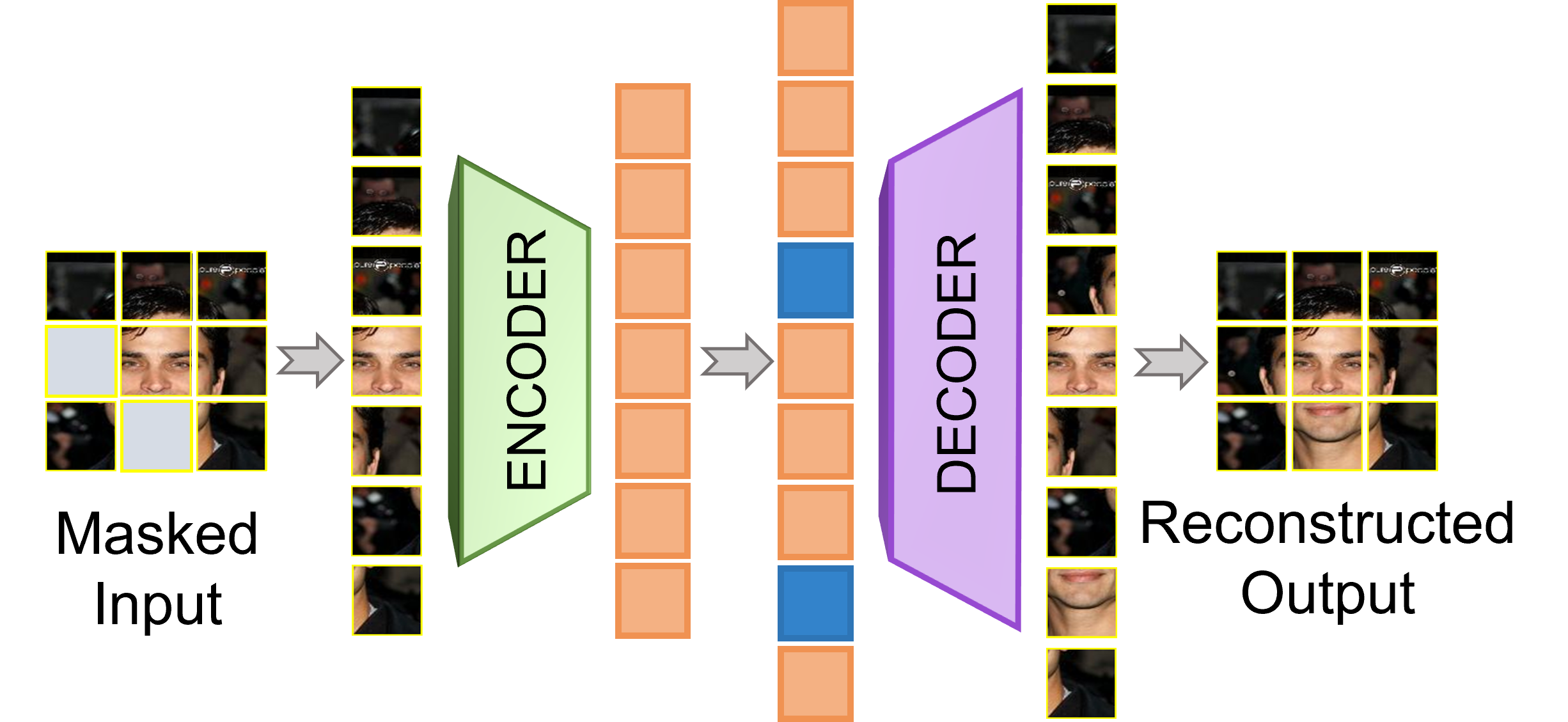}
   \caption{ViT based autoencoder architecture. During training, some parts of the input are masked and the visible patches are encoded and padded with zero tokens. Then, the decoder is used to reconstruct the masked patches.}
  \label{fig:mae}
\end{figure}

\subsection{Masking and Reconstruction Module}
We incorporate a masking and reconstruction module (MRM) to effectively capture appropriate syntax.~This choice is primarily motivated by two reasons.~First, we argue that a generative approach is more suited for syntactic understanding, as the DNN needs to learn proper spatial arrangements for generating meaningful images. Moreover, discriminative models (classifiers) fail to differentiate between syntactically CIs and IIs (Fig.~\ref{fig:motivation}), as the features of these images overlap in the hidden representational space (demonstrated later in Fig.~\ref{fig:pca}). Second, our approach is inspired by masked language modeling of BERT \cite{bert}, and GPT \cite{gpt}, which have resulted in LMs with syntactic understanding \cite{refining}. Hence, by intuition mimicking a similar masked LM approach for images would improve visual syntactic reasoning.

Our architecture is based on a ViT-based masked autoencoder, as displayed in Fig.~\ref{fig:mae}. Following ViT \cite{vit}, the input is divided into non-overlapping patches of 16 by 16 pixels. During training, we randomly mask 50\% of pixels and only the visible patches are encoded with a series of transformer blocks. Then, the mask tokens are padded and this whole set is processed through the decoder transformer blocks.~Unlike \cite{mae}, the decoder is used both during training and inference.~However, during inference, the masking is not random (different from \cite{mae}), rather the detected parts from the PD are sequentially masked one at a time and reconstructed.~This part-based masking while inferring is different from training. Training is performed with random masking so that the network not only learns to reconstruct the word locations but also any part of the input. This is particularly useful for images where parts (e.g.~ear or eye) are placed in random places (e.g.~eye in forehead or ear in background), and the network successfully erases these extra parts, despite never seeing such inputs in training.  

 \begin{figure*}[t]
  \centering
\includegraphics[width=0.95\linewidth]{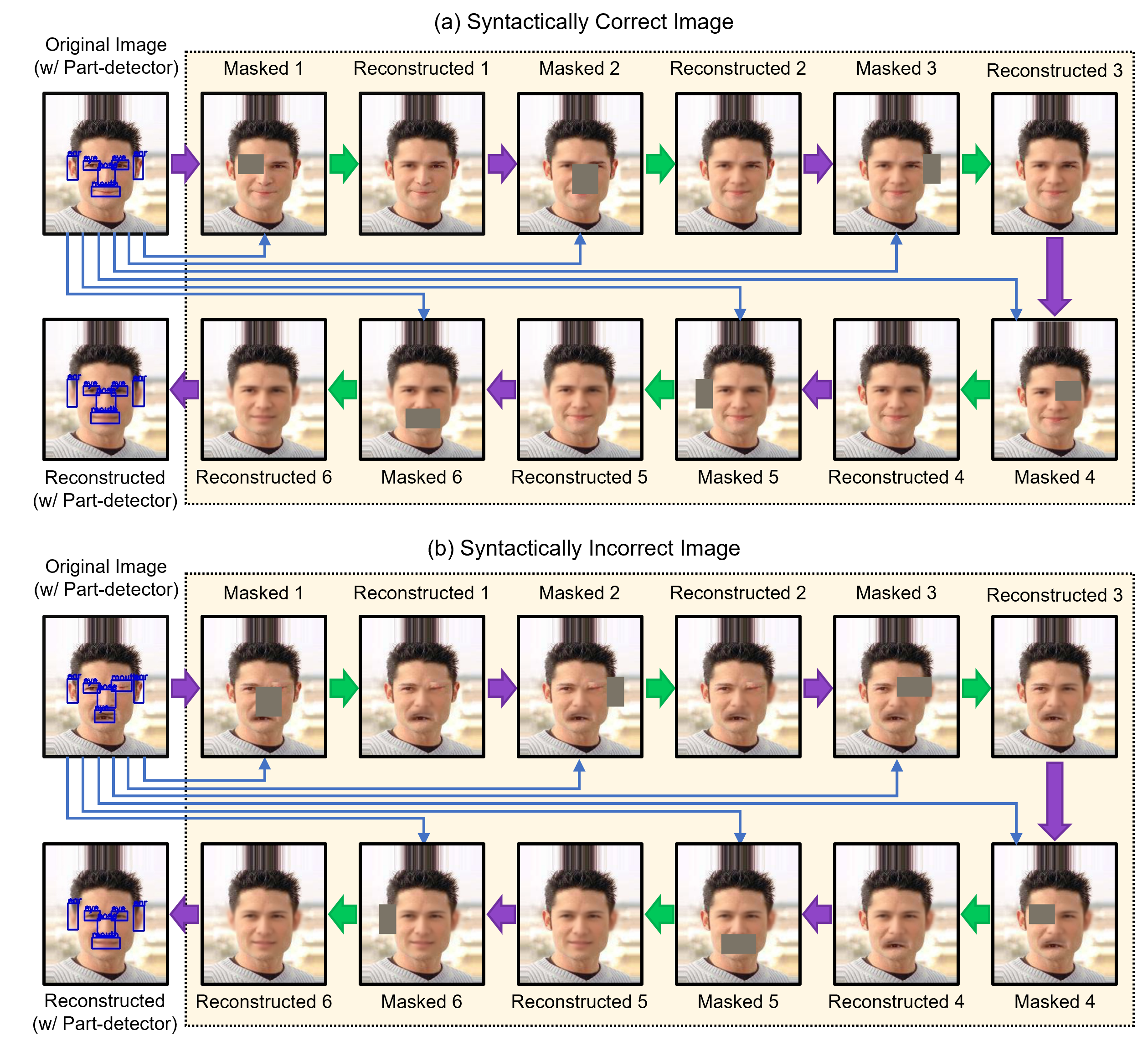}

   \caption{The reconstruction pipeline where the detected parts from the PD are masked sequentially and reconstructed using the autoencoder. Eventually, the output recovers the correct version of the input.}
  \label{fig:rec_pipeline}
\end{figure*}

A pipeline of the whole reconstruction process is shown in Fig.~\ref{fig:rec_pipeline}.~At each step, one word is masked and the network provides a plausible reconstruction at that location.~This process is repeated for all words present in the image sentence.~Note, from the second part onward, the reconstructed output from the previous step is used for masking instead of the original input. This enables reconstructing a corrected version of the given input in addition to syntax checking.~As shown in Fig.\ref{fig:rec_pipeline}(b), we can reconstruct a corrected face where the eye and mouth are swapped in input.~This process mostly leaves the syntactically CIs unchanged as expected (Fig.\ref{fig:rec_pipeline}(a)), since the reconstructions at each word location resemble the input.~The final reconstructed image is passed through the part detector to identify the words present in that image.                    

\begin{algorithm}[h]
   \caption{Pseudo-code of the syntax checker module}
   \label{alg:checker}
\begin{algorithmic}
   \STATE {\bfseries Input:}  Boxes of detected parts of the original ($B_{o}$) and reconstructed image ($B_{r}$) and their labels ($L_{o}$ and $L_{r}$), IOU threshold ($t$), classes corresponding to $L_{o}$ and $L_{r}$
   \STATE {\bfseries Initialize:} $n_{o}$ = length($B_{o}$), $n_{r}$ = length($B_{r}$), $c=1$
   
   \FOR{$i \leftarrow 1$ {\bfseries to} $n_{o}$}
          
        \STATE box1=$B_{o}[i]$, flag=0 
        \FOR{$j \leftarrow 1$ {\bfseries to} $n_{r}$}
                    \STATE box2=$B_{r}[j]$
                    \IF{IOU(box1,box2)$>t$}
                        \STATE flag=1
                        \IF{$L_{o}[i]!=L_{r}[j]$}
                            \STATE $c=0$ // part mismatch error
                            \STATE //~ Output- class($L_{o}[i]$) in place of class($L_{r}[j]$)
                        \ENDIF
                    \ENDIF

        \ENDFOR
        \IF{flag==0}
            \STATE $c=0$ // extra-part error
            \STATE //~ Output - class($L_{o}[i]$) in place of no specific part
        \ENDIF

   \ENDFOR
   \STATE \textbf{return} $c$ // $c=1$ denotes correct, $c=0$ denotes incorrect
\end{algorithmic}
\end{algorithm}

\subsection{Syntax Checker}
The detected words from the original input (O) and reconstruction (R) are processed using a syntax checker as shown in Fig.~\ref{fig:method} to determine correctness. The steps inside the checker block are depicted in Algorithm \ref{alg:checker}. For each detected word in O, we search if there is a word in R at the corresponding location. As the coordinates of R's words might not match exactly with O, this search is performed based on the intersection over union (IOU) between them.~If the part labels for all words of O match with R, syntax is deemed correct; whereas any mismatch denotes word swapping. Moreover, if a word of O does not match with any word of R, it indicates the presence of an extra word in O.     

\section{Experiments and Results}
\label{results}
\subsection{Datasets}
We evaluate our approach on CelebA \cite{liu2018large}, and AFHQ \cite{afhq} datasets. 2000 randomly selected images from CelebA are used for testing as CIs and the rest are used for training the MRM. AFHQ is divided into cat and wild. For these classes, we introduce 200 syntactically IIs from each class for testing. AFHQ has 5000 images each for cat and wild sets, of which 1000 from each category (chosen randomly) are used as correct testing data while the rest are used for training. The wild category has samples from tiger, cheetah, fox, wolf, and lion. Similarly, 200 randomly chosen images from Pascal-Part and ImageNet are edited to be tested as IIs alongside some synthetic samples. 
 \begin{figure*}[t]
  \centering
\includegraphics[width=0.95\linewidth]{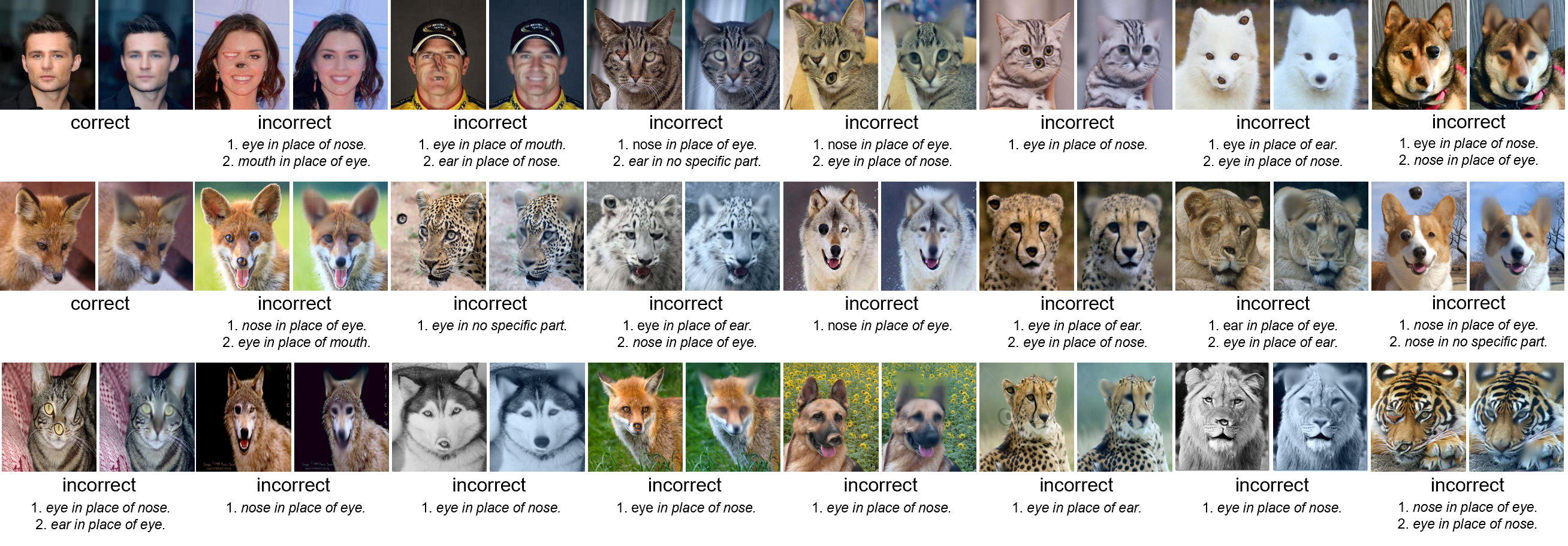}
   \caption{Visual results of the proposed method for correct as well as incorrect inputs from different classes.}
  \label{fig:visual_results}
\end{figure*}

 \begin{figure*}[t]
  \centering
\includegraphics[width=\linewidth]{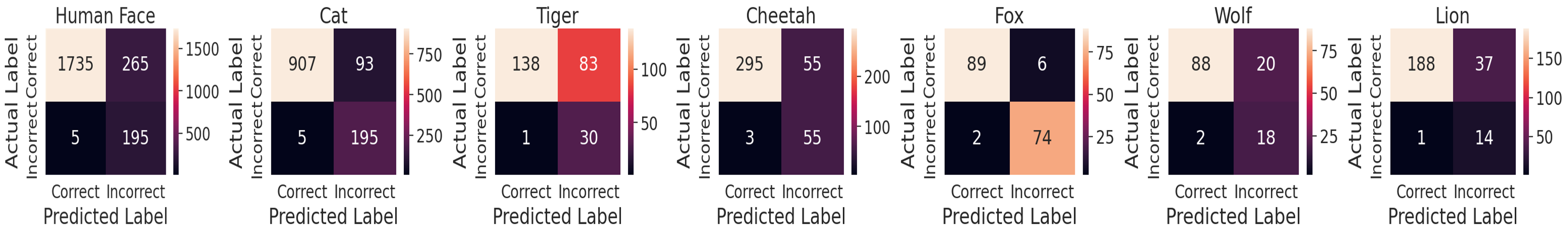}
   \caption{Confusion matrix of syntactic correctness prediction for the classes from CelebA, and AFHQ datasets.}
  \label{fig:conf_mat}
\end{figure*}

\subsection{Network architecture}
The part detector used for our experiments is trained by fine-tuning a faster-RCNN \cite{faster} having a ResNet-50 backbone with feature pyramid network. The masked autoencoder is based on ViT-base network \cite{vit}. The architectural details of the autoencoder are-

Encoder: embedding dimension-768, number of transformer blocks-12, number of attention heads-12, MLP ratio-4, normalization used- LayerNorm.

Decoder: embedding dimension-512, number of transformer blocks-8, number of attention heads-16, MLP ratio-4, normalization used- LayerNorm.

\subsection{Training Details}
\textbf{Part Detector.} 500 images from each class were manually labeled for training the part detector (450 used for training, 50 for validation). Note, the reconstructed images obtained after passing through the masked autoencoder are usually a bit blurry. As the part detector also needs to locate words on these reconstructed images, we augment our training set for the part detector with the reconstructed counterparts of the original training images for each class. Only correct images are used for training purposes. We train the part detector network with stochastic gradient descent optimization (weight decay=1e-6, momentum=0.9) to optimize for bounding box detection for the parts. We fine-tune the model for 30 epochs, with a batch size of 1 and an initial learning rate of 0.001. The learning rate is divided by 10 at every 5th epoch, till epoch 10. 

\textbf{Reconstruction Module.}
Standard data augmentation techniques are applied in this case such as cropping by randomly sampling from the padded image or its horizontally flipped version. Both training and testing data are normalized using channel-wise mean and standard deviation calculated from the ImageNet training set. The networks are trained with mean-squared error (MSE) loss using AdamW optimizer (weight decay=0.05, betas=0.9, 0.95). We train the models for 400 epochs for each dataset, with an initial learning rate of 0.001. We use mask ratio=0.5 during training, so 50\% of patches are randomly removed while training. Again, training uses the correct images only.   

\textbf{Syntax Checker.} For the syntax checker, the intersection over union (IOU) threshold was set to 0.3 for comparing corresponding box locations from the original input and reconstructed image. Additionally, for CelebA, the threshold for non-max suppression (NMS) was set to 0.1 and 0.3, for original input and reconstructed image, respectively. For the AFHQ dataset, this NMS threshold was set to 0.05 for both the original input and reconstructed image.

\subsection{Qualitative Results} In linguistics, 4 types of syntactic errors can be present- (i) swapping of words (e.g.~\textit{I person an am}), (ii) replacement with a wrong word (e.g.~\textit{I person a person}), (iii) extra word (e.g.~\textit{I am am a person}), (iv) omission of word (e.g.~\textit{I am a}). For visual syntax, we consider the first 3 types of errors, and their combinations. The omission of word is not considered because images can be syntactically correct despite some missing parts (due to crop, occlusion, pose etc.,~as shown in Fig.~\ref{fig:variation}).
~We present qualitative results using the proposed method in Fig.~\ref{fig:visual_results}. Each input is followed by its reconstruction.~For CIs, the parts remain similar after reconstruction.
Interestingly, for the IIs, the MRM successfully replaces the syntactically wrong parts with their expected counterparts. For example, for the 2nd image in row 1 of Fig.~\ref{fig:visual_results}, nose is replaced by an eye and an eye is replaced by mouth. The 4th image in that row (cat) has an extra ear at a random location.~Other IIs in Fig.~\ref{fig:visual_results} contain various syntactic anomalies. In row 3, we show results on some ImageNet samples from classes coinciding with the training set of CelebA and AFHQ. Our approach was able to correctly identify the corrupted images, thereby demonstrating the generalization efficacy of our method. 
Additional visual results with varying levels of difficulty are provided in Appendix Section A (Figs.~\ref{fig:results1}, \ref{fig:results2}, \ref{fig:results3}).

\subsection{Quantitative Results.}~Syntactically correct vs incorrect prediction is essentially a binary classification problem. Note our results represent an unsupervised approach as the incorrect images are never exposed during training. The number of IIs in our testing set is significantly lower compared to the correct ones, which is usual for any anomaly detection problem.~Therefore, we use balanced accuracy as a quantitative metric of classification performance taking the imbalance in the test set into account. For CelebA and AFHQ, we report the class-wise as well as overall metrics as the number of classes is smaller. The class-wise prediction performance for CelebA and AFHQ is shown using the confusion matrices in Fig.~\ref{fig:conf_mat}. They depict the count of true positive (TP), true negative (TN), false positive (FP), and false negative (FN) for each class. The formula for balanced accuracy is, Balanced Accuracy = (Sensitivity+Specificity)/2 \cite{balancedacc}, where Sensitivity = TP / (TP + FN) and Specificity = TN / (TN + FP). For the human face, cat, and wild, we obtain sensitivity scores of 0.867, 0.907, and 0.799, respectively. The specificity values for these 3 classes are 0.975, 0.975, and 0.955 respectively. Overall balanced accuracies obtained for face, cat, and wild are 92.10\%, 94.10\%, and 87.69\%, respectively. To further test for generalization, we experiment with 400 incorrect faces originally taken from the Caltech 101 face/motorbike dataset. Our model can identify the syntactically IIs with 96.27\% accuracy. 

\subsection{Interpretability}~Besides syntactic assessment, an additional desirable property of the proposed framework is interpretability. As shown in Figs. 4(b) and \ref{fig:visual_results}, the output can reconstruct incorrect parts with their correct counterparts and erase extra erroneous parts, highlighting the parts that led the model to predict the input as incorrect. Such interpretability would be unavailable for a simple binary DNN classifier. One of the major issues with DNNs is the blackbox nature of the operation, which our proposed technique attempts to counter to some extent. Even for the CIs, this interpretability factor is useful. Usually, if an input is classified as a face by a DNN, it is difficult to discern the reasoning behind the prediction. Essentially, the question remains \textit{what} gets a face \textit{classified} as a \textit{face?} With an end-to-end DL pipeline, realizing that is a challenge. However, as the proposed method decomposes the whole image into parts, we can reason that it is a face as it contains eyes, ears, nose, mouth etc.~in a proper syntax.

 \begin{figure}[t]
  \centering
  
\includegraphics[width=\linewidth]{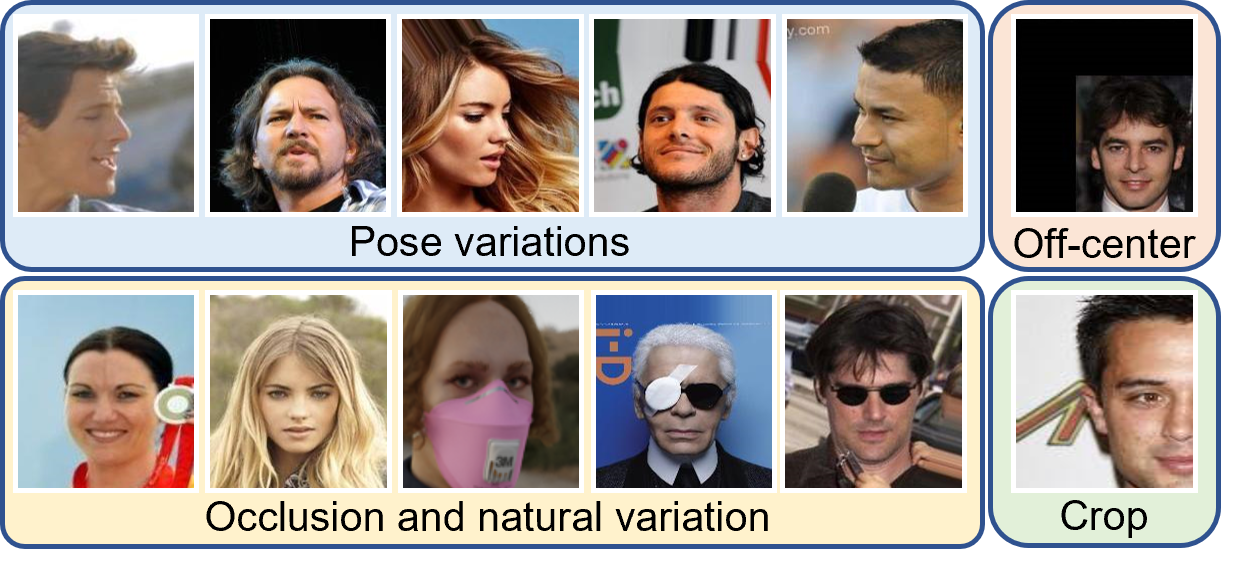}
   \caption{Example variations in the images of CelebA. Such variations make rule-based syntax checking infeasible.}
  \label{fig:variation}
\end{figure}

 \begin{figure}[t]
  \centering
\includegraphics[width=0.9\linewidth]{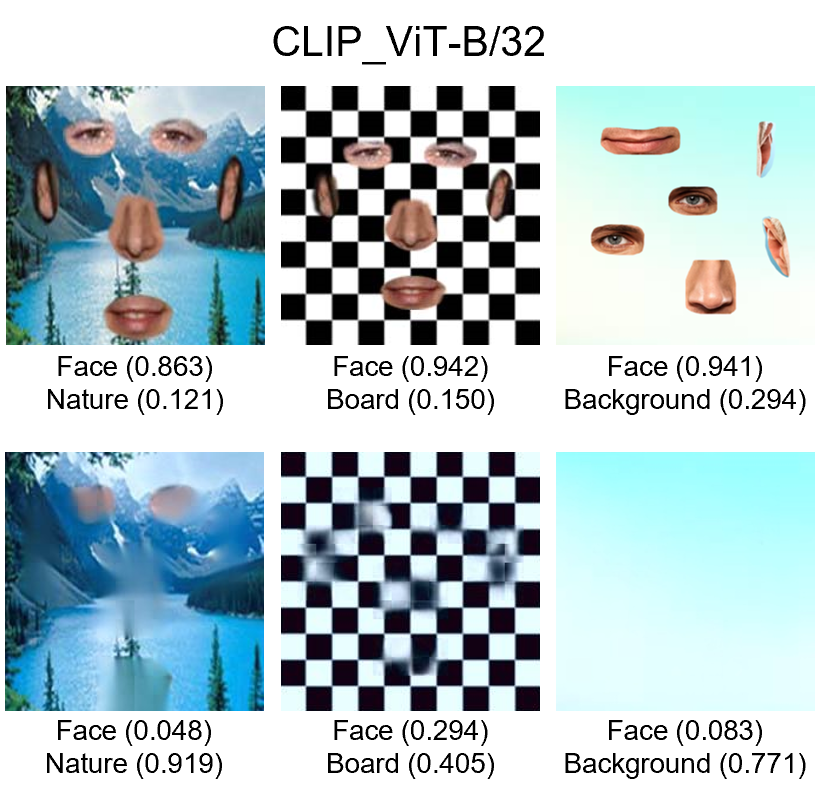}
   \caption{Face parts separated and scattered on natural scene (left), board (middle) and random background (right).~Reconstruction (bottom row) erases the parts, resulting in correct classification with CLIP$\textunderscore$ViT.}
  \label{fig:background}
\end{figure}

\begin{table}[t]
\small
\caption{OoD detection performance with a ResNet-18 trained on CelebA. Clearly, syntactically IIs are not OoD.}
\label{tab1}
\setlength{\tabcolsep}{5pt}
\begin{tabular}{ccc}\\ \hline 
OoD  & Detection Error @  & FPR @ \\
 Dataset & TPR 95\% &  TPR 95\% \\  \hline 
Syntactically Incorrect Faces	& 0.5070	&	0.9604 \\

Gaussian Noise	& 0.0	&	0.0 \\ 

Uniform Noise	& 0.0	&	0.0 \\ 

SVHN	& 0.0063 & 0.0011	 \\ 

CIFAR10	& 0.0547	&	0.0611 \\ 

CIFAR100	& 0.0603	&	0.0706 \\ 

Textures	& 0.0414	&	0.0351 \\ 

LSUN	& 0.1253	&	0.2017 \\ 

TinyImageNet	& 0.0707	&	0.0953 \\ 

Places365	& 0.0561	&	0.0672 \\ \hline

\end{tabular}
\end{table}

\subsection{Significance of Different Modules} In this section, we perform an ablation study on the importance of the 3 salient blocks in the proposed method.~First, we analyze if it is feasible to perform syntactic evaluation without the MRM. With just the PD, the syntax checking has to be performed based on a set of if-then-else rules.~However, PD followed by rule-based checking does not lead to a robust scheme since the number of combinations leading to syntactic anomalies is exponentially huge. Hence, capturing all the combinations is not possible.~Moreover, there would be no learning involved in this framework.~Even for CIs, the possible configurations can vary largely.~Due to pose variations, only certain parts may be visible; some parts may be occluded or images may be cropped as depicted in Fig.~\ref{fig:variation}. Furthermore, an image may have varying aspect ratios leading to different distances among parts like the off-center picture in Fig.~\ref{fig:variation}. Similarly, if face parts are separated and scattered on random backgrounds (Fig.~\ref{fig:background}), PD followed by a rule-based check detects it as an image with correct syntax.~Even a CLIP$\textunderscore$ViT-B/32 model is fooled with such inputs.~These are a few representative cases only (with innumerable similar possibilities) where a rule-based syntax checker without the MRM would fail. Notably, with the proposed PD and MRM pipeline, we are able to rectify this issue, as shown in Fig.~\ref{fig:background}.~Next, we investigate the performance of MRM in isolation.~In absence of the PD, the masking is performed randomly.~We tried with mask ratios of 0.1 and 0.25.~For both cases, CIs were reconstructed using the trained model and we recorded the maximum mean squared error (MSE) of reconstruction.~Later, we used this maximum MSE as threshold for syntactic evaluation of the test set (MSE higher than the threshold indicates anomaly).~Interestingly, we obtained only 50\% accuracy, which is a chance level performance for a 2-class scenario.~Hence, MRM in isolation is not sufficient for detecting erroneous syntax.~Lastly, the syntax checker block is obviously critical as it performs the eventual decision-making.~Without it, checking just based on the reconstruction error gives random results (50\% accuracy).~Furthermore, the reconstruction error-based checking does not provide interpretation of incorrectness, which the proposed IOU-based checker module offers.   

 \begin{figure}[t]
  \centering
\includegraphics[width=\linewidth]{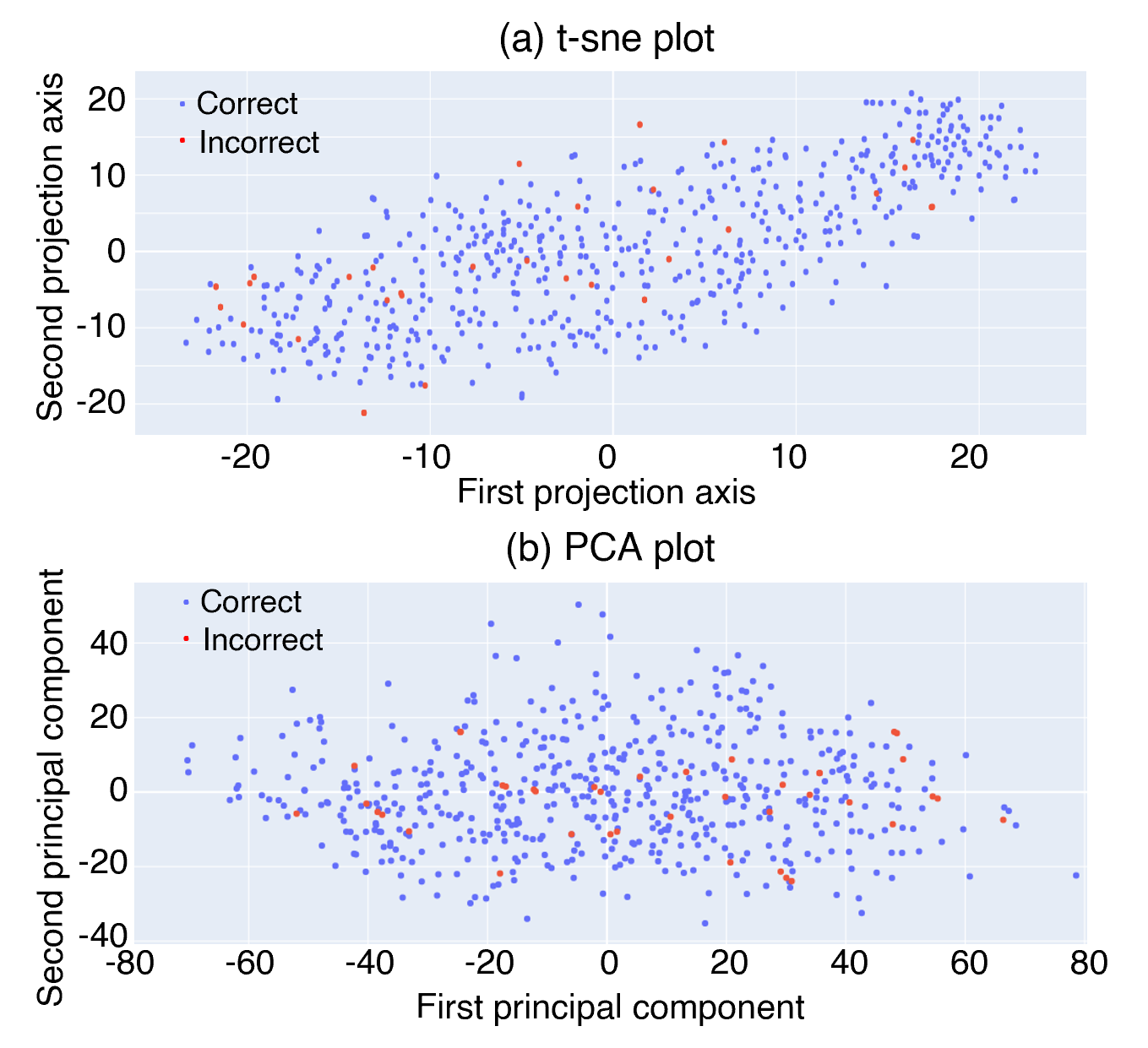}
   \caption{Visualization of feature space using (a) t-sne, (b) PCA. The syntactically correct and incorrect images overlap in the hidden states, thereby limiting their separability.}
  \label{fig:pca}
\end{figure}

\begin{table*}[t]
\small
\caption{Comparison of the proposed method to other detection algorithms on syntactically correct vs incorrect faces.}
\label{tab2}
\setlength{\tabcolsep}{5pt}
\begin{tabular}{cc|cc|cc}\\ \hline
Method  & Accuracy(\%) & Method  & Accuracy(\%) & Method  & Accuracy(\%) \\ \hline

Binary classifier	& 50.0	 & OoD detector \cite{lee2018simple}& 50.0 &	AnoGAN \cite{anogan}	& 51.7 \\

ClipCap \cite{clipcap}	& 50.0 & Raw-OC-SVM \cite{scholkopf1999support}	&	50.0 &	Simplenet \cite{liu2023simplenet}	& 57.3 \\ 

KDE \cite{kde}	& 50.0	& CAE-OC-SVM \cite{scholkopf1999support}&	56.7 &	Normalizing flows \cite{chiu2023self}	& 54.7 \\ 

Transformations \cite{izak}	& 50.0  & DFM \cite{dfm}  & 52.9	&	Attention-augmentations \cite{bozorgtabar2023attention}	& 52.4  \\ 
DSEBM \cite{dsebm}	& 50.4	& DAGMM \cite{dagmm} &	51.2 &	Omni-frequency
channel-selection \cite{liang2023omni}	& 51.8 \\ 

Ganomaly \cite{ganomaly}	& 50.7 & CAE \cite{masci2011stacked}	&	50.0 &	Diffusion models \cite{zhang2023unsupervised}	& 56.2 \\ 

ADGAN \cite{ADGAN}	& 52.6 &	Deep SVDD \cite{svdd} &	53.3 &	\textbf{Ours} &	\textbf{92.1} \\ \hline

\end{tabular}
\end{table*}

\subsection{Are Samples with Incorrect Syntax OoD?} Next, we investigate if out-of-distribution (OoD) detection can identify IIs using the well-studied Mahalanobis OoD detector \cite{lee2018simple}.~First, we train a ResNet-18 on CelebA to predict the 40 attributes of a face image.~Then, the Mahalanobis OoD detector is used to obtain the detection performance metrics to separate various datasets and the II faces. From the results given in Table \ref{tab1}, we see that the detection error and false positive rate (FPR) values of the II faces are the same as random chance, while we can detect other OoD datasets with very low error.~This demonstrates that syntactic anomaly identification is not an OoD detection problem. Additionally, we visualize the features of syntactically CIs and IIs. We train on CelebA images using a convolutional autoencoder with 5 conv layers in the encoder and 5 transposed conv layers in the decoder.~After training, the encoded features corresponding to syntactically CIs and IIs are obtained and we perform t-sne \cite{van2008visualizing} and principal component analysis (PCA) on them.~The results are elucidated in Fig.~\ref{fig:pca}. Interestingly, both plots show that CIs and IIs overlap considerably in the hidden states, further validating that separating them is inherently challenging.

\subsection{Binary Classification} As a sanity check, we explore if a binary classifier can perform  CI versus II classification. To this end, we fine-tune a ResNet-18 pre-trained on ImageNet for binary classification on the CelebA faces. 160 out of the 200 incorrect images from our set of IIs were used for training and the rest for testing. After convergence of training, we find the balanced accuracy to be 50\% on the test set, which is again at the level of random guessing.~Remarkably, none of the IIs were correctly identified.~This occurs due to the feature overlap between the 2 classes as described earlier (Fig.~\ref{fig:pca}). Even if the binary classification approach were to work for this case, it would not be a feasible solution since- (i) we require labeled data for the incorrect class, which is not feasible as the possible incorrect configurations for each correct case are theoretically uncountable, 
(ii) a binary classifier does not provide any interpretation of prediction. Note, however, that for visual syntactic anomalies, such binary classification does not even work in the first place.   

\subsection{Comparative Performance}
\label{comparison}
We compare the balanced classification performance of various methods on syntactically CIs versus IIs and provide the results in Table \ref{tab2}.~The experiments with binary classification and OoD detection have been described earlier.~We also try with an image captioning model, Clipcap \cite{clipcap}, to capture syntax through captions. However, these approaches completely fail to identify the IIs. Additionally, we experimented with scene graph-based approaches \cite{johnson2015image,xu2017scene} to capture relationships among various image parts. However upon training with CIs only, this method failed to
infer with IIs. Next, we investigate the applicability of anomaly detection methods on syntax checking.~One-class SVM \cite{scholkopf1999support} and deep SVDD \cite{svdd} methods perform slightly better than chance level, but the accuracy is not satisfactory. We also try with other anomaly detection methods, such as GAN-based \cite{anogan,ADGAN,ganomaly}, kernel density estimation (KDE) \cite{kde}, energy-based model \cite{dsebm} and Gaussian mixture model \cite{dagmm}.~These approaches mainly attempt to model the distribution of the normal class and anomalies are detected based on their large distance from the distribution of normal samples in the feature space. However, from the results in Table \ref{tab2}, these methods are clearly not suited for visual syntactic anomaly detection.~We attribute this behavior to the fact that IIs in our case do not come from a different distribution compared to CIs, as demonstrated in Table \ref{tab1} and Fig.~\ref{fig:pca}. Additionally, we experimented with various latest anomaly detection methods including attention-based \cite{bozorgtabar2023attention} approach, normalizing flows \cite{chiu2023self}, frequency-based method \cite{liang2023omni}, and diffusion models \cite{zhang2023unsupervised}, however they are ineffective for the CI versus II detection task, as illustrated in Table \ref{tab2}.  Furthermore, the anomaly detection methods \cite{izak} assume that the anomalous samples belong to a different class than the normal class (in a cat versus dog problem, the cat is the normal class while the dog is the anomaly class); whereas, our syntactic anomalous samples originate from the same underlying class. Therefore, the problem considered here is considerably more challenging compared to usual anomaly detection. 

\begin{figure}[t]
  \centering
   \includegraphics[width=\linewidth]{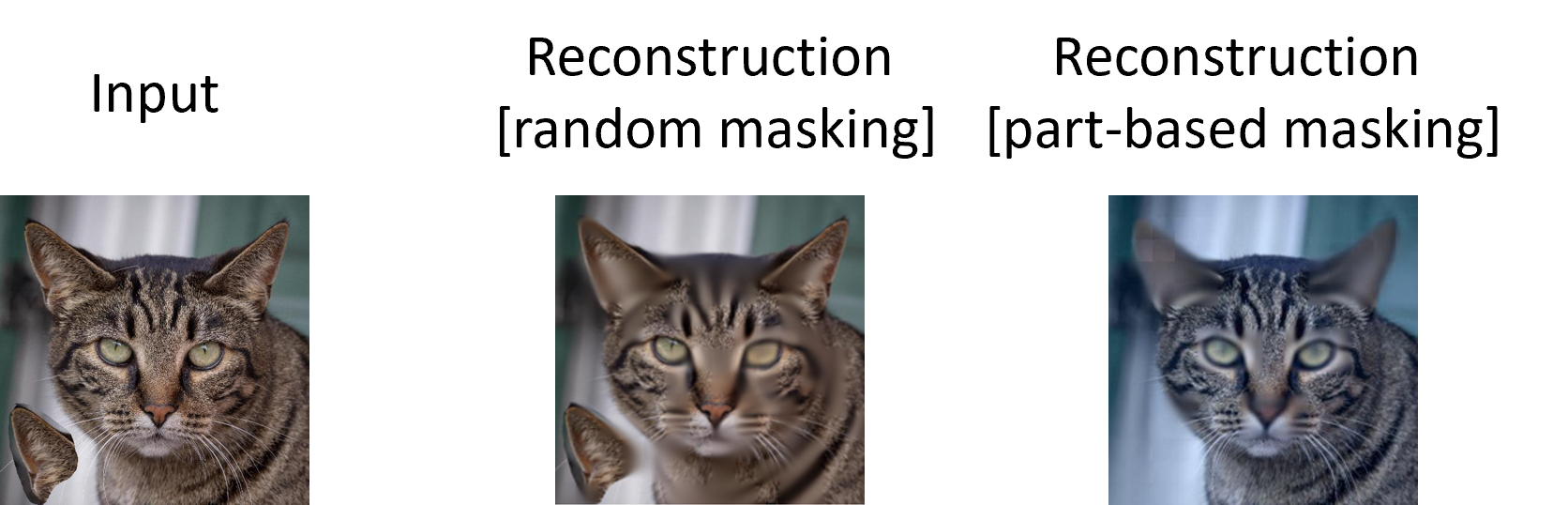}
   \caption{Comparison between our method (part-based masking) and random masking.}
  \label{fig:fig18}
\end{figure}

\subsection{Random masking versus Part-based masking} 
We validate our choice of part-based masking during inference over random masking in this section. Random masking often results in missing extra parts as shown in Fig.~\ref{fig:fig18}, or might miss misplaced parts (wrong syntax). As a result, the MRM module is unable to rectify those through reconstruction and the syntax checker ends up giving wrong predictions. However, as shown in Fig.~\ref{fig:fig18}, our method is successful in handling these cases (removing the extra ear in this case) and provides the correct output. Overall, the CI versus II detection accuracy drops to 62.38\% from 92.1\% (with part-based masking) on CelebA, demonstrating the significance of the proposed approach.

\subsection{Analysis of Challenging Failure Cases}
\begin{figure}[t]
  \centering
\includegraphics[width=\linewidth]{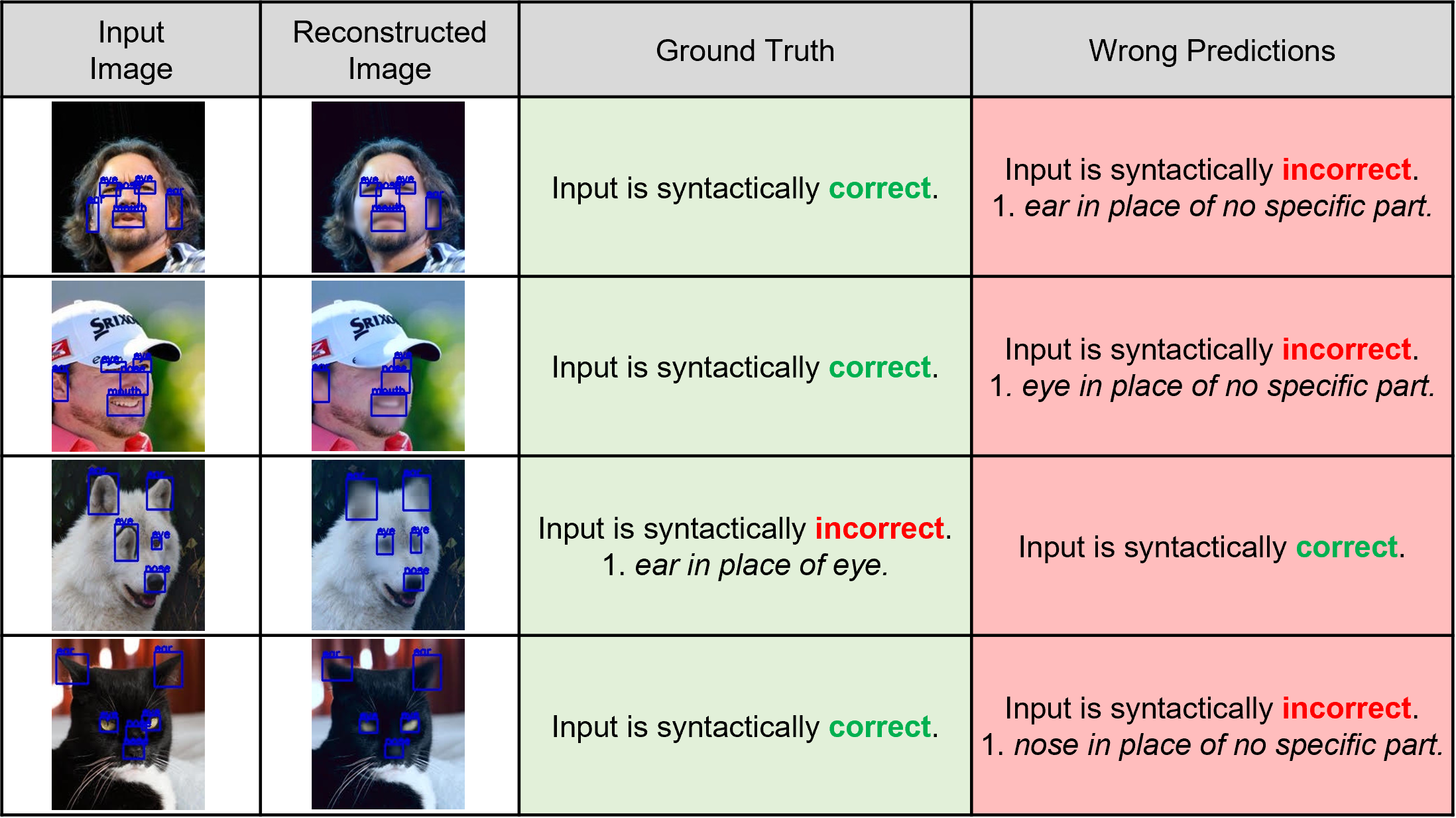}
   \caption{Failure analysis on some samples with erroneous prediction.}
  \label{fig:failurecases}
\end{figure}

We analyze some example challenging failure cases in this section as shown in Fig.~\ref{fig:failurecases}. For the image in the 1st row, the left ear is hardly visible due to hair. As a result, although the initial part detector can detect it correctly, the reconstruction module almost wipes it out. So, this ear is not detected as a word post reconstruction, which leads to an error in prediction. Similarly, for the image in the 2nd row, the left eye is almost occluded by the hat, so the reconstruction module wipes it out. Consequently, this eye is no longer detected after reconstruction and the decision is erroneous. These are some challenging cases. However, sometimes error occurs for relatively easy inputs too as can be seen for the 3rd row image. Here, the part detector mistakenly detects the ear (placed in the left eye) as an eye. Another sample difficult case for a cat image is shown on the fourth row, where above the actual nose, the part detector erroneously detects an extra nose (due to the presence of a dark nose-like shape). As a result, for these last 2 cases, although the reconstruction is successful, our output is wrong. So, the error can originate from both the part detector as well as the reconstruction module.  

\begin{figure}[t]
  \centering
\includegraphics[width=\linewidth]{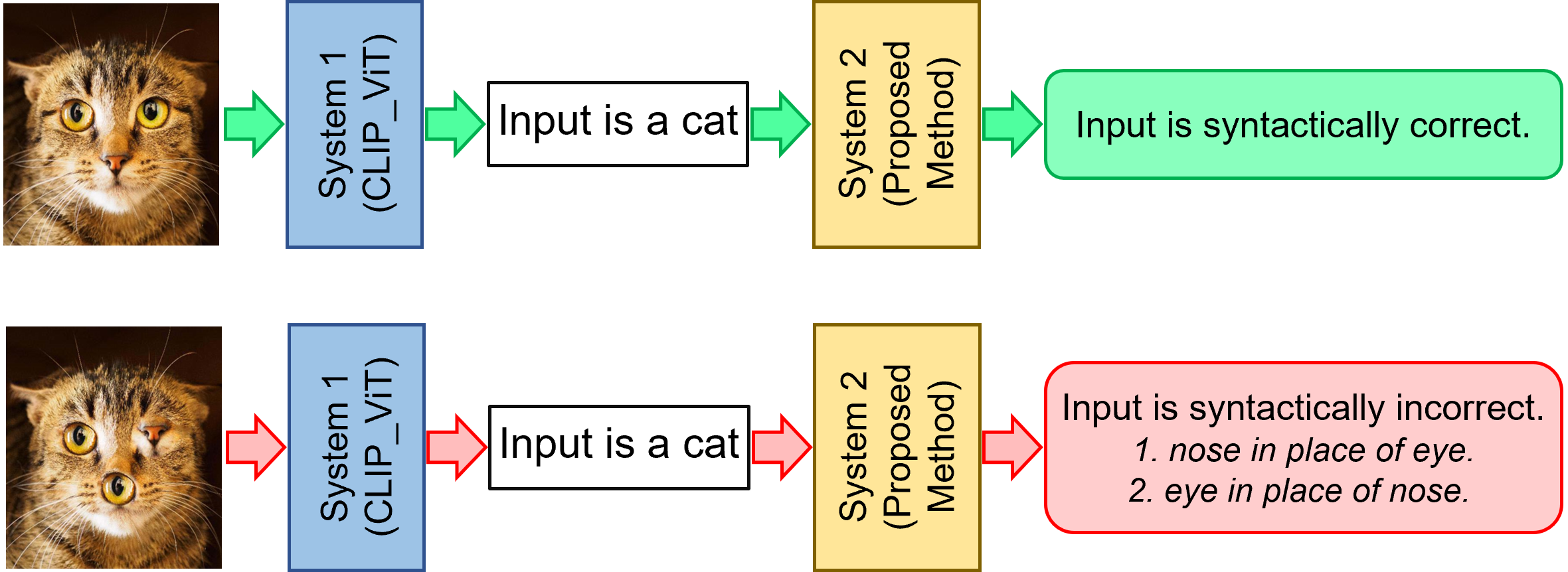}
   \caption{Schematic of the proposed method as system 2 of a neuro-symbolic pipeline. System 1 (a CLIP$\textunderscore$ViT model) executes the initial broad classification and subsequently, system 2 performs fine-grained syntactic evaluation.}
  \label{fig:sys12}
\end{figure}

\subsection{Neuro-Symbolic Angle}~A two-stage pipeline (System-1 and System-2) for cognitive processing has been proposed in  \cite{kahneman2011thinking}.~System 1 performs fast, automatic pattern recognition, while system 2 executes slower, step-by-step deliberation \cite{nsai}. In this regard, DL is considered as a system 1 process and rule-based approaches are part of system 2 \cite{neurosymbolic}.~Aligning with this theme, our proposed approach can also be viewed from a neuro-symbolic lens as depicted in Fig.~\ref{fig:sys12}.~We consider CLIP$\textunderscore$ViT \cite{clip} DL model as system 1 and our proposed method as an example of system 2. As shown in Fig.~\ref{fig:sys12}, system 1 performs a fast broad classification.~Hence, for example, a syntactically correct as well as an incorrect cat are both predicted similarly as a cat. Following this, our proposed approach (as a system 2) executes the step-by-step syntactic evaluation. From a neuro-symbolic point of view, we may consider the neural part to comprise of the CLIP$\textunderscore$ViT as well as our part detector and MRM blocks; whereas the syntax checker module represents the symbolic part.

\section{Conclusion}
\label{conclusion}
Syntactic reasoning, in addition to semantics, is a pivotal aspect of holistic scene understanding. While NLP models such as BERT have been shown to possess considerable syntactic understanding, in this article, we report current DNN-based computer vision models usually struggle to capture the visual syntax. This phenomenon is a novel Achilles heel unveiled through our work. Furthermore, to remedy this issue, we take inspiration from the masked autoencoding technique (used in pre-training of BERT) and introduce a 3-stage framework combining part detection and sequential masking with reconstruction followed by a checker module. The proposed language-model motivated approach obtains 92.10\%, and 90.89\% accuracy on syntactic correctness evaluation, on CelebA, and AFHQ datasets, respectively.~Through analysis of hidden state features from the trained models, we observe that syntactically correct and incorrect images overlap significantly in this latent space.~As a result, current OoD or anomaly detection algorithms are unable to perform the desired syntactic evaluation, whereas the proposed approach provides satisfactory performance.  Some potential real-world applications of our work include: industrial scenarios- detecting if objects are in the required specific spatial configuration (production line), content moderation detection- checking if the content has been modified from expectation, anomaly detection- to check if spatial anomaly exists in industrial products (e.g. chips). A limitation of the proposed solution is that it has not been tested on very large-scale datasets. However, note that since the problem itself is novel, no specific large-scale dataset for syntactic evaluation exists and the main motivation of this work stems from an understanding point of view. We reveal an intriguing failure mode of DNNs concerning visual syntax to draw the attention of the community. Future works will involve exploring more complex syntactic relationships and large-scale datasets, experimenting with other types of language models, etc. To conclude, we believe that enabling visual syntactic understanding within a DL pipeline is critical towards achieving a truly intelligent NN-based agent.~Our work takes a small step in that direction by exposing an existing problem in current DNNs and by providing a potential solution combining DNN-based neural blocks and a syntax checker module.

\appendices
\section{ Additional Qualitative Results}
 We provide additional visual results in Figs.~13-15. Both correct and incorrect cases for all classes are displayed. These results demonstrate the efficacy of the proposed pipeline.

\section*{Acknowledgments}
This work was supported in part by the Center for Co-design of Cognitive Systems (CoCoSys), one of the seven centers in JUMP 2.0, a Semiconductor Research Corporation (SRC) program sponsored by DARPA, by the SRC, the National Science Foundation, Intel Corporation, the DoD Vannevar Bush Fellowship, and by the U.S. Army Research Laboratory.


 \begin{figure*}[t]
  \centering
\includegraphics[width=0.9\linewidth, height=9.5in]{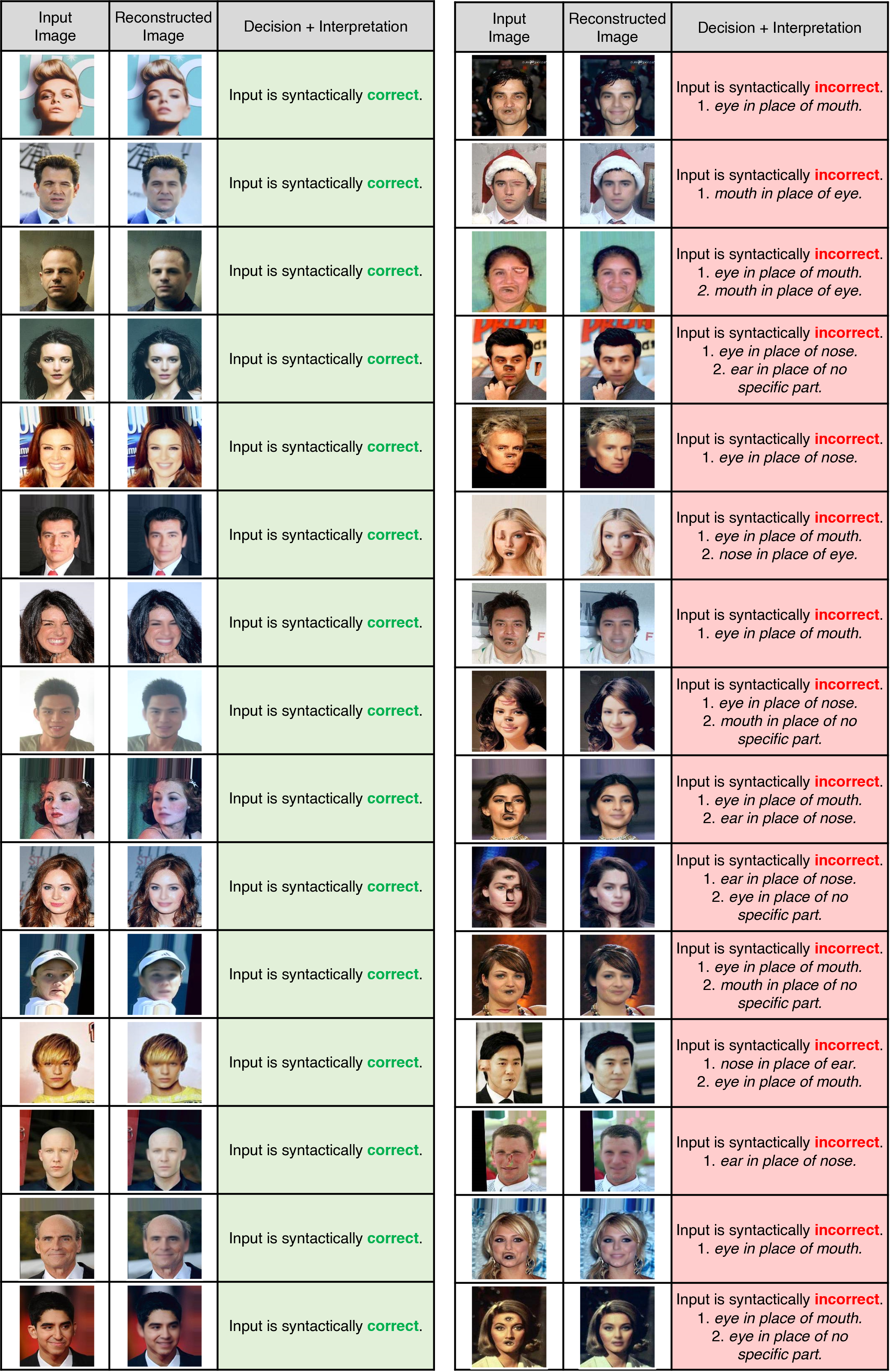}
   \caption{Additional qualitative results (A), with syntactically correct images on the left and incorrect images on the right.}
  \label{fig:results1}
\end{figure*}

 \begin{figure*}[t]
  \centering
\includegraphics[width=0.9\linewidth, height=9.5in]{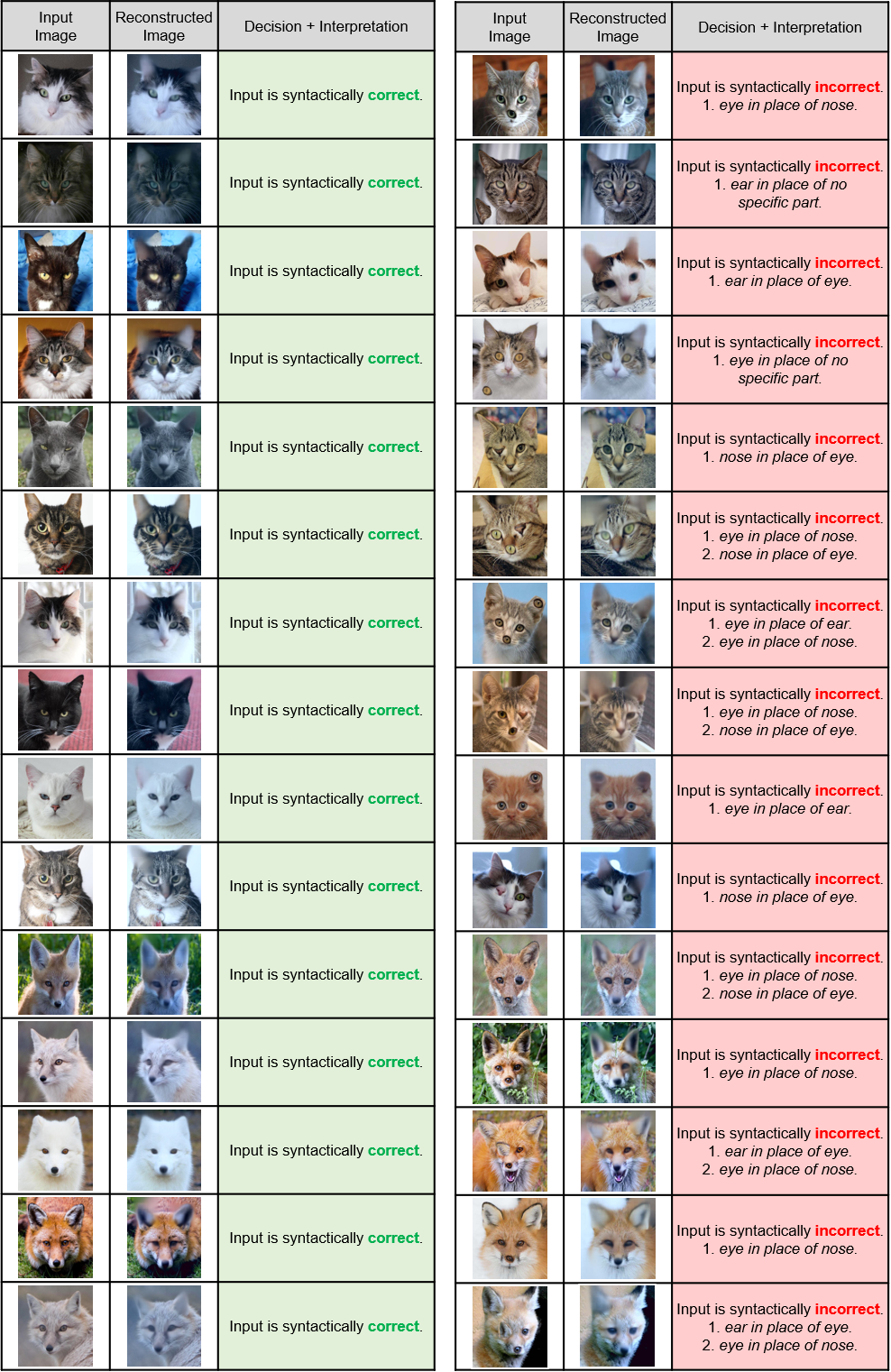}
   \caption{Additional qualitative results (B), with syntactically correct images on the left and incorrect images on the right.}
  \label{fig:results2}
\end{figure*}

 \begin{figure*}[t]
  \centering
\includegraphics[width=0.9\linewidth, height=9.5in]{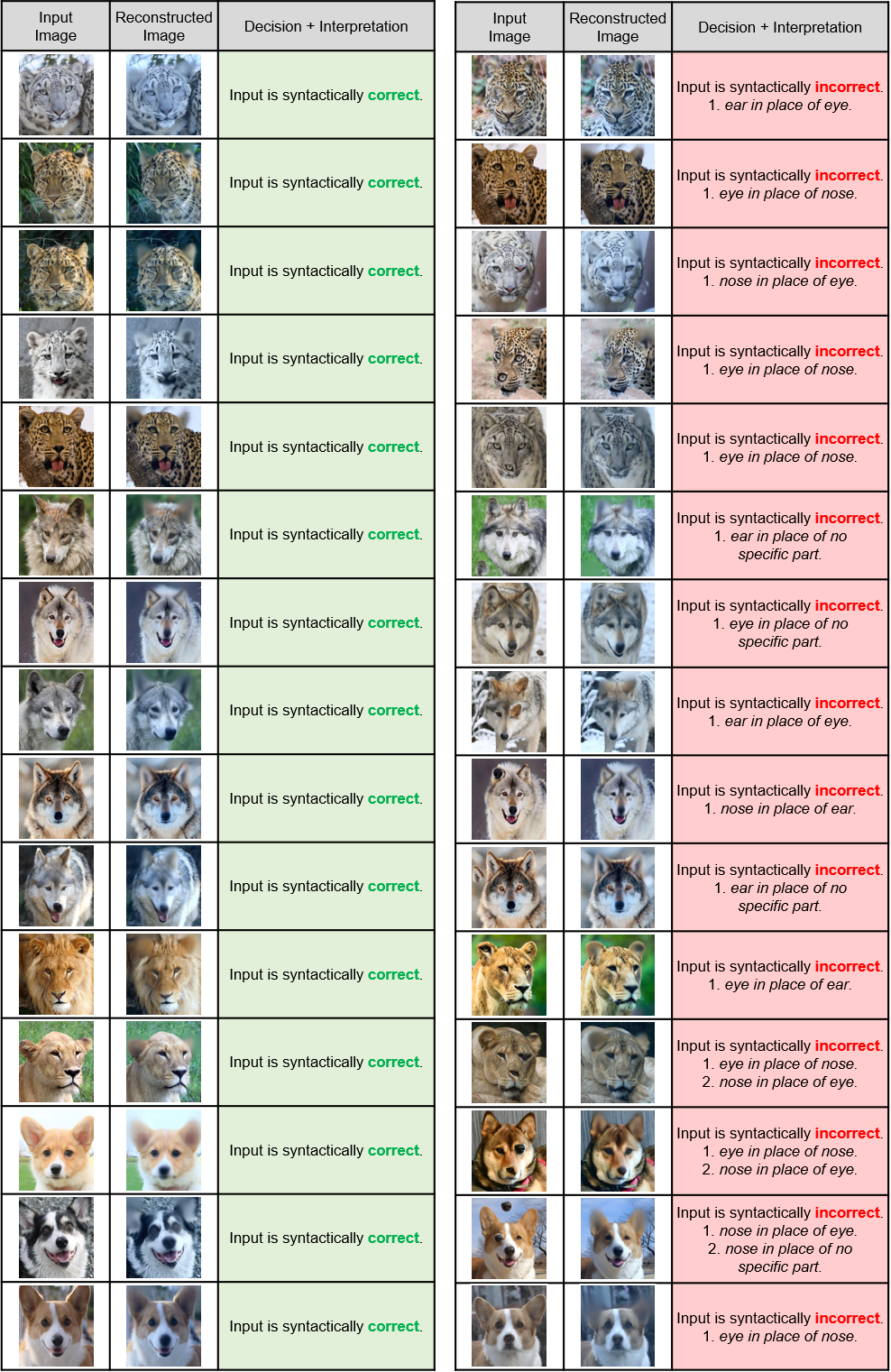}
   \caption{Additional qualitative results (C), with syntactically correct images on the left and incorrect images on the right.}
  \label{fig:results3}
\end{figure*}

\bibliography{sample}

\end{document}